\documentclass[letterpaper]{article} 
\usepackage[preprint]{aaai2027}  
\usepackage[hyphens]{url}  
\usepackage{graphicx} 
\usepackage{natbib}  
\usepackage{caption} 
\usepackage{newfloat}
\usepackage{listings}
\DeclareCaptionStyle{ruled}{labelfont=normalfont,labelsep=colon,strut=off} 
\DeclareFloatingEnvironment[fileext=lst,name=Listing,placement=tb]{listing}

\usepackage{booktabs}
\usepackage{amsmath}
\usepackage{multirow}
\usepackage{array}
\usepackage{xcolor}

\renewcommand{\textfraction}{0.05}

\renewcommand{\dbltopfraction}{0.95}
\renewcommand{\dblfloatpagefraction}{0.80}

\newcommand{\best}[1]{\textbf{#1}}
\newcommand{\winner}[1]{\textbf{\underline{#1}}}
\DeclareMathSizes{7.5}{7}{5}{5}

\definecolor{codebg}{HTML}{F7F8FA}
\definecolor{codeframe}{HTML}{D9DEE8}
\definecolor{codeblue}{HTML}{1F5AA6}
\definecolor{codegreen}{HTML}{287A4D}
\definecolor{codepurple}{HTML}{7A3E9D}
\lstdefinelanguage{Typst}{
  morekeywords={import,let,set,show,if,else,for,in},
  morecomment=[l]{//},
  morestring=[b]",
  sensitive=true
}
\lstdefinestyle{seaslidescode}{
  basicstyle=\footnotesize\ttfamily,
  numbers=none,
  backgroundcolor=\color{codebg},
  frame=single,
  rulecolor=\color{codeframe},
  framesep=5pt,
  xleftmargin=2pt,
  xrightmargin=2pt,
  aboveskip=3pt,
  belowskip=3pt,
  columns=fullflexible,
  keepspaces=true,
  breaklines=true,
  keywordstyle=\color{codeblue}\bfseries,
  commentstyle=\color{codegreen},
  stringstyle=\color{codepurple},
  showstringspaces=false
}
\lstdefinestyle{seaslidesprompt}{
  basicstyle=\fontsize{7.5pt}{8.2pt}\selectfont\ttfamily,
  numbers=none,
  backgroundcolor=\color{codebg},
  frame=single,
  rulecolor=\color{codeframe},
  framesep=4pt,
  xleftmargin=2pt,
  xrightmargin=2pt,
  aboveskip=3pt,
  belowskip=4pt,
  columns=fullflexible,
  keepspaces=true,
  breaklines=true,
  breakatwhitespace=true,
  breakindent=0pt,
  breakautoindent=false,
  showstringspaces=false
}

\title{SeaSlides: Semantic Abstraction Layer for Agentic Slide Generation}

\author{
    Shengjun Fang\textsuperscript{\rm 1,\rm 2},
    Chenyang Wu\textsuperscript{\rm 1,\rm 2},
    Zongzhang Zhang\textsuperscript{\rm 1,\rm 2}\corresponding
}
\affiliations{
    \textsuperscript{\rm 1}National Key Laboratory for Novel Software Technology, Nanjing University, China\\
    \textsuperscript{\rm 2}School of Artificial Intelligence, Nanjing University, China\\
    fangsj@lamda.nju.edu.cn, wucy@lamda.nju.edu.cn, zzzhang@nju.edu.cn
}

\begin{document}

\maketitle

\begin{abstract}
Agentic presentation generation must preserve source content, maintain coherent visual design, render specialized objects, and produce usable artifacts. Existing systems meet only part of this requirement: templates preserve regularity but restrict adaptation, whereas free-form HTML or SVG gives models flexibility at the cost of low-level rendering decisions. This mismatch makes long technical decks brittle, especially when slides contain formulas, code, or data graphics. We present \textbf{SeaSlides}, an agentic slide-generation framework built around \textbf{a semantic abstraction layer}. Rather than authoring coordinates, inline styles, or raw SVG geometry, the model writes structured slide content through reusable components and capability modules, while templates own layout, style, and rendering. We instantiate this principle separately in HTML and Typst: SeaSlides-HTML uses template-defined DOM components, whereas SeaSlides-Typst uses template functions and package-backed modules. Capability modules route equations, code, and charts to dedicated renderers, and three feedback stages localize build errors, project-constraint violations, and visual defects before export. The two systems retain backend-specific syntax and contracts while sharing the same authoring boundary. For evaluation, we combine the 128-task UltraPresent validation setting with \textbf{SeaSlidesBench-Rich}, a new 32-task benchmark stressing mathematics, code, pseudocode, tables, charts, and diagrams. Across four generation models, both SeaSlides backends produce more readable, content-oriented source than SVG-heavy generation. A SeaSlides backend attains the highest rich-content macro-average under three of the four models while maintaining competitive overall qualitative performance. These results support semantic abstraction as a practical authoring principle across presentation backends.
\end{abstract}

\section{Introduction}

Presentations are a primary medium for communicating technical ideas in research, education, and professional practice. Generating one requires more than summarizing a document. A system must preserve source meaning across a multi-slide narrative, produce a coherent visual artifact, and leave that artifact editable after export. Technical decks add a further constraint because equations, code, and data graphics depend on specialized rendering rather than ordinary text layout.

Recent systems seek more expressive representations for this task. PPTAgent edits PowerPoint structures with reference-slide guidance~\citep{zheng2025pptagent}; AutoPresent generates structured visual programs~\citep{ge2025autopresent}; PPT-Master uses SVG as an editable intermediate~\citep{he2026pptmaster}; and DeepPresenter generates HTML slides and revises them from rendered observations~\citep{zheng2026deeppresenter}. These systems show the value of executable artifacts, but also expose a persistent trade-off. Rigid templates preserve visual regularity while limiting adaptation. Free-form code, SVG, and HTML offer greater control, but make the model responsible for more of the visual implementation.

The underlying problem is that the output representation often doubles as a rendering-level authoring interface. Coordinate programs require direct placement decisions, and SVG requires geometry and styling. HTML can rely on browser layout, but generated files still tend to intermingle slide content with the surrounding visual system and one-off corrections. The model must therefore preserve source facts while designing each slide, then debug the representation it has just produced. Across a long deck, style drifts and visual logic is duplicated. Global revision becomes harder, while mathematics, code, and other technical objects are often approximated instead of delegated to reliable renderers.

We argue that presentation agents should instead work through a \emph{semantic abstraction layer}. The model chooses the narrative and evidence, then expresses each slide through operations declared by a template contract. The backend owns the layout, styling, and specialized rendering behind those operations. This is not fixed-template filling: the contract constrains the implementation vocabulary without fixing the content, slide sequence, or component choices.

\begin{figure*}[t]
    \centering
    \includegraphics[width=\textwidth]{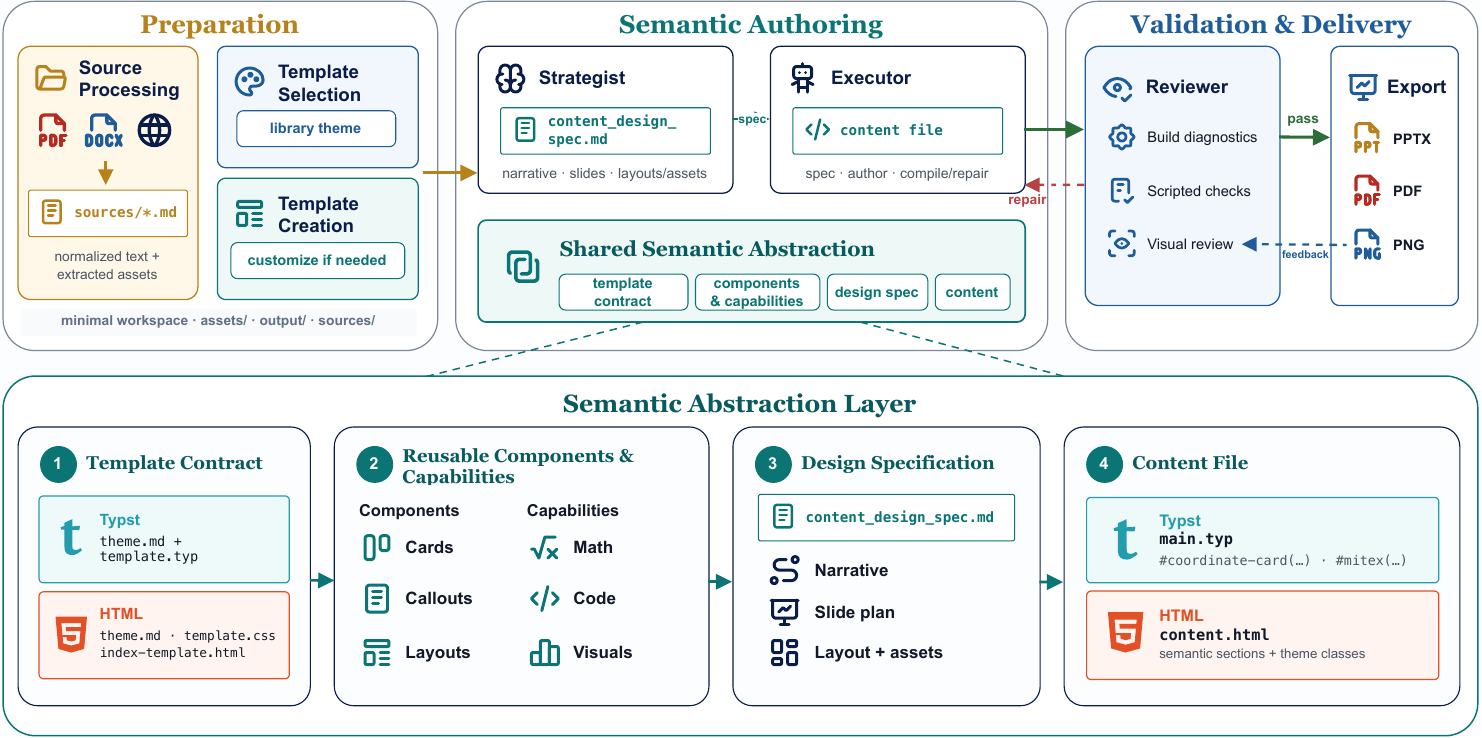}
    \caption{SeaSlides workflow and its semantic abstraction layer. Preparation supplies source material and a selected or newly created template; the Strategist records narrative and design intent in \texttt{content\_design\_spec.md}, which the Executor turns into backend-specific semantic content. Staged review returns actionable failures to the Executor before export.}
    \label{fig:overview}
\end{figure*}

To instantiate this idea, we introduce \textbf{SeaSlides}, an agentic framework for reliable and editable slide generation. Figure~\ref{fig:overview} shows the division of work. The Strategist turns the request, source material, and template contract into a deck-level design specification; the Executor realizes that plan as backend-specific semantic content. The content file carries source-grounded statements, slide structure, and speaker notes, while the template defines typography and component behavior. Reusable components capture recurring presentation structures, and capability modules render objects that benefit from dedicated tools.

SeaSlides couples this authoring layer with staged environment feedback. Build diagnostics first expose local execution errors, scripted checks then test project-level requirements, and visual review is reserved for defects that require rendered evidence. The ordering provides a specific repair signal and avoids image-based review for failures already localized by the environment.

We realize the same principle in two backend-specific systems. \textbf{SeaSlides-HTML} combines template-defined DOM components with Chromium rendering, using MathJax and Prism for specialized content before serializing the settled slide DOM through SVG for PNG and editable PPTX export. \textbf{SeaSlides-Typst} uses Typst~\citep{typst2023} and Touying~\citep{touying2024}, compiling PDF and PNG directly and using an SVG intermediate for editable PPTX. The two systems retain different syntax, contracts, and rendering stacks; what they share is the responsibility boundary between semantic authoring and visual implementation.

We evaluate on the 128-task UltraPresent validation setting and the new 32-task \textbf{SeaSlidesBench-Rich}. Both SeaSlides backends produce substantially more readable, content-oriented source than SVG-heavy generation. On rich-content tasks, a SeaSlides backend leads under three of four generation models while remaining competitive overall.

Our main contributions are threefold:
\begin{enumerate}
    \item We propose \textbf{a semantic abstraction layer} as a model-facing authoring principle for agentic slide generation. The model writes structured content through reusable operations instead of directly producing rendering geometry and local styling.
    \item We realize this layer in \textbf{SeaSlides-HTML} and \textbf{SeaSlides-Typst}, each with its own component contract and rendering path, and combine it with staged environment feedback.
    \item We introduce \textbf{SeaSlidesBench-Rich}, a 32-task benchmark for technical presentations, and evaluate it alongside the 128-task UltraPresent validation setting using both rendered-quality and model-authored-source metrics.
\end{enumerate}

\section{Related Work}

\paragraph{Document-to-slide generation.}
Earlier systems largely cast slide generation as content selection followed by layout. PPSGen extracts salient sentences from academic papers~\citep{ppsgen2013}; DOC2PPT jointly models content and layout for scientific documents~\citep{fu2022doc2ppt}; and D2S retrieves text, figures, and tables before query-conditioned summarization~\citep{sun2021d2s}. Later systems decompose long-document generation into outline construction, content grounding, and slide realization~\citep{bandyopadhyay2024enhancing,liang2025slidegen}. This line of work establishes the importance of source fidelity and deck-level narrative. SeaSlides addresses a different layer of the problem: the representation through which an agent expresses that narrative to a renderer.

\paragraph{Templates, editing, and visual programs.}
Presentation systems differ in how much visual implementation they expose to the model. KCTV fills manually designed templates~\citep{cachola2024kctv}, while PPTAgent selects and edits reference slides~\citep{zheng2025pptagent}. AutoPresent generates Python visual programs with SlidesLib helpers~\citep{ge2025autopresent}; PPT-Master and DeepPresenter instead use SVG and HTML as executable representations~\citep{he2026pptmaster,zheng2026deeppresenter}. These choices trade a constrained editing space for progressively greater control over layout and rendering.

Other systems combine markup generation with multimodal review or interactive editing~\citep{xu2025pregenie,yang2025autoslides}, train agents for presentation aesthetics~\citep{liu2026evopresent}, or manipulate PowerPoint objects and learned style preferences~\citep{jung2025talk,zeng2025slidetailor}. Beamer, Typst, and Touying likewise demonstrate the value of declarative authoring outside the agent setting~\citep{beamer2023,typst2023,touying2024}. SeaSlides differs in where it places the authoring boundary: the agent writes semantic content through a backend contract, while the backend retains the visual implementation.

\paragraph{Agentic refinement and verification.}
Execution feedback is now common in tool-using agents~\citep{yao2023react,wu2023autogen,hong2024metagpt}. Self-Refine uses model-generated feedback~\citep{madaan2023selfrefine}, whereas CRITIC grounds correction in external tools~\citep{gou2024critic}. Presentation and visual-program agents likewise revise failed edit actions, rendered artifacts, or executable code~\citep{zheng2025pptagent,zheng2026deeppresenter,si2025design2code,yang2024matplotagent}. SeaSlides does not claim rendered observation itself as new. Its distinction is to order heterogeneous signals by cost and locality: textual build failures first, deterministic project checks second, and rendered-slide inspection only for defects that require visual evidence.

\paragraph{Presentation benchmarks and evaluation.}
AutoPresent introduces SlidesBench with reference-based and reference-free slide metrics~\citep{ge2025autopresent}; PPTAgent proposes PPTEval for content, design, and deck coherence~\citep{zheng2025pptagent}; and DeepPresenter evaluates constraint satisfaction, content, and style on UltraPresent~\citep{zheng2026deeppresenter}. PPTBench targets layout understanding and manipulation~\citep{huang2025pptbench}, while PresentBench uses instance-specific binary rubrics for real-world deck evaluation~\citep{chen2026presentbench}. These efforts follow a broader shift toward rubric-guided LLM evaluation~\citep{liu2023geval}. Representation quality nevertheless remains underexplored: a plausible slide can still be backed by brittle source, and standard tasks rarely stress equations, code, pseudocode, or dense technical diagrams. Our evaluation retains the UltraPresent validation tasks for comparability and adds SeaSlidesBench-Rich together with source- and artifact-oriented measurements.

\section{SeaSlides}

\subsection{Workflow Overview}

Figure~\ref{fig:overview} organizes SeaSlides into three stages. \textbf{Preparation} processes optional sources, selects or creates a backend template, and initializes its workspace. During \textbf{semantic authoring}, the Strategist records deck-level intent in \texttt{content\_design\_spec.md}; the Executor turns that specification into backend-specific content. \textbf{Validation and delivery} then runs the Reviewer on compiled output, returns localized failures to the Executor, and exports the repaired deck. The middle stage is mediated by the semantic abstraction layer described next: HTML and Typst share its division of responsibilities, but not a source language.

\subsection{Semantic Abstraction Layer}

The lower panel of Figure~\ref{fig:overview} expands the four artifacts that mediate semantic authoring. On the backend side, the \emph{template contract} declares the available operations, which reusable components and capability modules implement. On the agent side, the \emph{design specification} records deck-level decisions and the \emph{content file} invokes those operations. We call the interface between these two sides the \emph{semantic abstraction layer}. SeaSlides-HTML and SeaSlides-Typst instantiate this layer separately through their own syntax and build paths; the four artifacts are parts of the interface, not serial processing layers.

The Strategist combines the request and normalized source material with the template contract. Its output is a design specification that connects the deck narrative to slide-level evidence and presentation intent. The Executor reads this specification under the same contract, maps each planned slide to supported operations, and writes the content file. Thus the Strategist decides \emph{what to communicate and how to organize it}; the Executor decides \emph{how to express that plan through the selected backend}.

For either backend \(b\in\{\mathrm{HTML},\mathrm{Typst}\}\), this handoff and the resulting responsibility boundary can be summarized as
\[
\begin{aligned}
d &= S(q,x,\kappa_b),\\
c_b &= E(d,\kappa_b),\\
y_b &= R_b(c_b;\theta_b,\mathcal{M}_b).
\end{aligned}
\]
Here \(q\) is the request, \(x\) contains normalized sources and assets, and \(\kappa_b\) is the backend's template contract. The Strategist \(S\) produces the design specification \(d\); the Executor \(E\) produces the semantic content file \(c_b\); and the renderer \(R_b\) combines that file with the template implementation \(\theta_b\) and capability modules \(\mathcal{M}_b\) to obtain the rendered deck \(y_b\).

This factorization also assigns repairs to explicit owners. Errors in narrative or evidence revise \(d\) or \(c_b\); recurring visual defects revise \(\theta_b\); and renderer-level failures in equations, code, or charts are handled in \(\mathcal{M}_b\). The Reviewer can therefore return localized evidence without asking the Executor to rewrite the full slide representation.

\subsection{Template-Content Separation}

Template-content separation (TCS) enforces the authoring boundary at the file level. The model-authored content file records slide structure and invokes semantic operations; the template owns their visual definitions and runtime dependencies. SeaSlides-HTML therefore separates \texttt{content.html} from \texttt{template.css} and \texttt{index-template.html}, while SeaSlides-Typst separates \texttt{main.typ} from \texttt{template.typ}.

Crucially, TCS constrains what the model authors, not merely where files are stored. The model may choose a comparison or algorithm component, but it does not recreate card styling, color rules, or absolute positions. A global design change therefore remains a template edit, and the content source can be inspected without tracing local rendering code. The matched appendix examples show the same slide intent in both backends; neither content file specifies coordinates, font sizes, or component internals.

\subsection{Components and Capability Modules}

Components and capability modules provide two forms of reuse. Components package recurring presentation structures such as comparisons, timelines, and theorem slides. Capability modules delegate objects whose correctness depends on a specialized renderer, including mathematics, code, and data graphics. Guided by the Strategist's specification, the Executor parameterizes these operations instead of recreating their implementation slide by slide.

The HTML backend implements components as DOM structures and CSS classes. It uses MathJax for equations, Prism for code, and browser APIs for data graphics; before SVG serialization, the compiler waits for these renderers and inlines local images.

The Typst backend implements the same distinction through functions and template rules. Native math and package-backed utilities handle specialized content, while compiler diagnostics localize syntax and type errors before visual review.

\subsection{Staged Environment Feedback}

\begin{figure}[t]
    \centering
    \includegraphics[width=\linewidth]{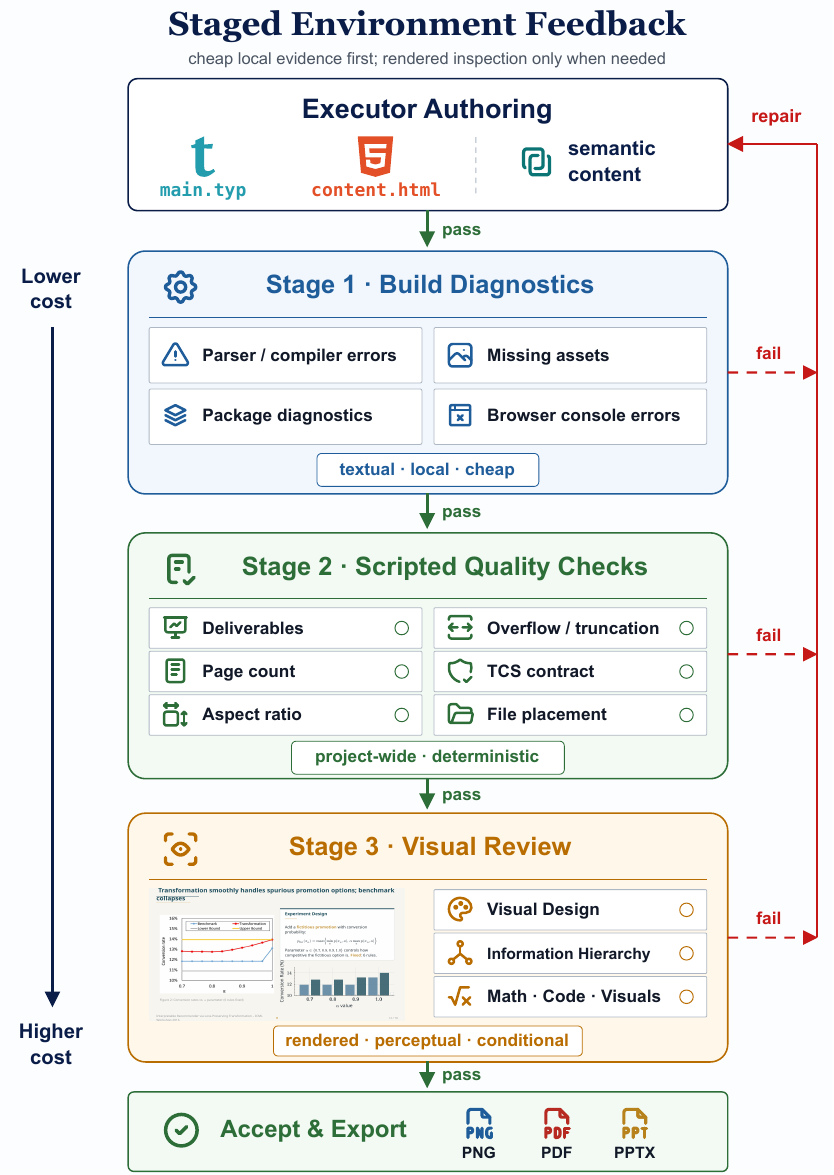}
    \caption{SeaSlides stages feedback by cost and locality: local build diagnostics first, deterministic project checks second, and visual review only after lower-cost checks pass.}
    \label{fig:feedback-workflow}
\end{figure}

Once the Executor produces executable content, three feedback stages validate it in increasing order of cost (Figure~\ref{fig:feedback-workflow}). They are validation stages, not additional layers in the authoring interface.

\paragraph{Stage 1: build diagnostics.}
The first signal comes from the backend itself: parser or compiler diagnostics, browser errors, and unresolved assets. These signals are textual, local, and cheap.

\paragraph{Stage 2: scripted quality checks.}
After compilation, deterministic scripts check artifact completeness and task-level constraints. They also detect structural failures such as blank pages, overflow symptoms, and TCS violations that compilation alone does not expose.

\paragraph{Stage 3: visual review.}
Only after the previous stages pass does the agent inspect rendered slides. Visual review targets hierarchy, contrast, and technical rendering defects that require perceptual evidence.

A failed stage returns diagnostics to the Executor; later stages run only after earlier failures are repaired. This ordering preserves the most specific evidence and avoids visual review for errors already localized by parsers, compilers, or scripts.

\subsection{Export Workflow}

The backends retain distinct source and rendering paths. SeaSlides-HTML serializes each slide DOM to SVG, then rasterizes it to PNG or converts it to editable PPTX. SeaSlides-Typst compiles PDF and PNG directly; editable PPTX export compiles Typst to SVG before converting vectors to PowerPoint objects. The content files are not interchangeable; what transfers is the division of responsibility between model-authored content and backend-owned rendering.

\section{Evaluation Tasks and Metrics}

\paragraph{Tasks.}
Our 160-task evaluation setting combines two named task sets. The first is the 128-task UltraPresent validation setting used by DeepPresenter~\citep{zheng2026deeppresenter}, which enables direct comparison on standard topic-only and source-conditioned generation. The second is our \textbf{SeaSlidesBench-Rich}, 32 additional tasks numbered 129--160 that stress technical objects and long source documents. Its tasks cover mathematical notation, code or pseudocode, and data graphics, often in combination. The dataset appendix summarizes both sets and the overlapping capability tags of SeaSlidesBench-Rich.

\paragraph{Systems and models.}
We compare KCTV~\citep{cachola2024kctv}, DeepPresenter~\citep{zheng2026deeppresenter}, PPT-Master~\citep{he2026pptmaster}, SeaSlides-HTML, and SeaSlides-Typst. Each system uses four generation models: Claude Sonnet 4.6, Gemini 3.1 Pro, Qwen 3.6 Plus, and KIMI K2.5.

\paragraph{Programmatic metrics.}
Constraint satisfaction checks the explicit presentation-level requirements that are programmatically observable: page count, language, and aspect ratio when specified by a task. Source length is reported in thousand characters over the relevant generated source. Content ratio measures the fraction of LLM-generated source devoted to content rather than style. Copied templates and generated artifacts are excluded from the main denominator.

\paragraph{LLM/VLM metrics.}
GPT-5.5 evaluates all automated LLM/VLM metrics on a 1--5 integer scale. Source readability is text-only: the evaluator inspects the model-authored source, not slide images. Content quality and visual style are evaluated from rendered slide images. On the tagged subset of tasks 129--160, we separately evaluate mathematical rendering, code rendering, and visualization. The three rubrics respectively assess mathematical semantics and typesetting, code/pseudocode correctness and presentation, and the data and structural fidelity of charts, tables, and diagrams. Appendix~\ref{app:prompts} reports the shared evaluator template and all six metric-specific prompts.

\paragraph{Diversity.}
Diversity uses DINOv2 features~\citep{oquab2024dinov2} and Vendi Score~\citep{friedman2023vendi}. The main table uses the first content slide policy to avoid title slides and section dividers dominating the estimate. Other policies are recorded by the evaluation script for auditing.
Unless otherwise noted, aggregate cells report mean$\pm$standard deviation across evaluated tasks or projects. Diversity is a group-level score and is therefore reported without a task-level standard deviation. In the main system comparison, higher-is-better metrics are ranked within each generation model, excluding DeepPresenter: the best value is bold and underlined, and other values within 0.05 of the best are bold. For source length, values below 15K characters are bold, and the lowest value for each model is also underlined.

\section{Experiments}

We organize the experiments around four questions: how semantic authoring affects output and source quality (RQ1), how faithfully the systems render technical content (RQ2), how template-content separation and each feedback stage affect the system (RQ3), and how the design behaves when the agent must create a template (RQ4).

\subsection{RQ1: Overall Quality and Representation}

Table~\ref{tab:main} reports the full system comparison. DeepPresenter is included for context but excluded from winner highlighting because its scores use completed-output subsets of 37, 39, 49, and 83 presentations across the four models; the other systems share the same 160 tasks. Appendix Table~\ref{tab:deeppresenter-results} reports both all-run and completed-output summaries.

Across all four models, the clearest difference from PPT-Master, the SVG-heavy baseline, lies in source readability and content ratio. Both SeaSlides backends score substantially higher on these representation-oriented metrics while retaining competitive rendered quality. With Claude Sonnet 4.6, their average qualitative scores are 4.07$\pm$0.44 and 4.06$\pm$0.33, compared with 3.57$\pm$0.43 for PPT-Master and 3.14$\pm$0.39 for KCTV.

\begin{table*}[!t]
\centering
\fontsize{8pt}{9pt}\selectfont
\setlength{\tabcolsep}{0.8pt}
\renewcommand{\arraystretch}{1.00}
\begin{tabular*}{\textwidth}{@{\extracolsep{\fill}}llcccc@{\hspace{4pt}\vrule\hspace{4pt}}ccrc@{}}
\toprule
\multicolumn{1}{c}{System} &
\multicolumn{1}{c}{Model} &
\multicolumn{1}{c}{Content} &
\multicolumn{1}{c}{Style} &
\multicolumn{1}{c}{Readability} &
\multicolumn{1}{c@{\hspace{4pt}\vrule\hspace{4pt}}}{Average} &
\multicolumn{1}{c}{Constraint} &
\multicolumn{1}{c}{CR} &
\multicolumn{1}{c}{Length} &
\multicolumn{1}{c@{}}{Diversity} \\
\midrule
\multirow{4}{*}{KCTV} & Claude Sonnet 4.6 & 2.57$\pm$0.89 & 2.89$\pm$0.31 & 3.97$\pm$0.51 & 3.14$\pm$0.39 & 0.505$\pm$0.228 & 0.470$\pm$0.169 & \winner{3.56$\pm$1.37}\phantom{0} & 0.249 \\
 & Gemini 3.1 Pro & 2.63$\pm$0.94 & 2.50$\pm$0.55 & 3.89$\pm$0.53 & 3.00$\pm$0.40 & 0.505$\pm$0.250 & 0.476$\pm$0.165 & \winner{3.66$\pm$1.50}\phantom{0} & 0.238 \\
 & Qwen 3.6 Plus & 2.61$\pm$0.90 & 2.52$\pm$0.51 & 3.73$\pm$0.56 & 2.95$\pm$0.43 & 0.500$\pm$0.234 & 0.472$\pm$0.165 & \winner{3.54$\pm$1.43}\phantom{0} & 0.244 \\
 & KIMI K2.5 & 2.61$\pm$0.83 & 2.62$\pm$0.52 & 3.81$\pm$0.51 & 3.01$\pm$0.39 & 0.507$\pm$0.237 & 0.468$\pm$0.168 & \winner{3.55$\pm$1.40}\phantom{0} & 0.252 \\
\midrule
\multirow{4}{*}{DeepPresenter} & Claude Sonnet 4.6 & 3.16$\pm$0.69 & 2.70$\pm$0.85 & 3.38$\pm$0.59 & 3.08$\pm$0.46 & 0.959$\pm$0.120 & 0.196$\pm$0.118 & 29.06$\pm$21.24 & 0.505 \\
 & Gemini 3.1 Pro & 3.28$\pm$0.83 & 2.69$\pm$0.80 & 3.36$\pm$0.74 & 3.11$\pm$0.58 & 0.936$\pm$0.156 & 0.184$\pm$0.094 & 29.28$\pm$18.39 & 0.520 \\
 & Qwen 3.6 Plus & 3.43$\pm$0.84 & 2.88$\pm$0.73 & 3.33$\pm$0.83 & 3.21$\pm$0.51 & 0.959$\pm$0.110 & 0.198$\pm$0.088 & 27.81$\pm$17.34 & 0.543 \\
 & KIMI K2.5 & 4.01$\pm$0.80 & 3.55$\pm$0.72 & 3.20$\pm$0.66 & 3.59$\pm$0.51 & 0.962$\pm$0.111 & 0.181$\pm$0.090 & 38.85$\pm$20.84 & 0.538 \\
\midrule
\multirow{4}{*}{PPT-Master} & Claude Sonnet 4.6 & \best{4.18$\pm$0.90} & \winner{4.28$\pm$0.65} & 2.26$\pm$0.57 & 3.57$\pm$0.43 & 0.873$\pm$0.189 & 0.079$\pm$0.035 & 127.81$\pm$92.32 & 0.452 \\
 & Gemini 3.1 Pro & 2.92$\pm$1.19 & 3.12$\pm$0.87 & 2.19$\pm$0.55 & 2.74$\pm$0.65 & 0.828$\pm$0.254 & 0.106$\pm$0.051 & 24.01$\pm$22.90 & 0.521 \\
 & Qwen 3.6 Plus & \winner{3.93$\pm$0.97} & \winner{3.86$\pm$0.54} & 2.18$\pm$0.44 & 3.32$\pm$0.41 & 0.871$\pm$0.207 & 0.079$\pm$0.032 & 85.54$\pm$58.86 & 0.453 \\
 & KIMI K2.5 & \winner{3.61$\pm$1.26} & \winner{3.79$\pm$0.62} & 2.15$\pm$0.60 & 3.18$\pm$0.56 & 0.846$\pm$0.235 & 0.076$\pm$0.036 & 82.06$\pm$40.16 & 0.421 \\
\midrule
\multirow{4}{*}{SeaSlides-HTML} & Claude Sonnet 4.6 & \winner{4.21$\pm$0.83} & 4.04$\pm$0.80 & 3.96$\pm$0.21 & \winner{4.07$\pm$0.44} & \winner{0.981$\pm$0.077} & 0.262$\pm$0.086 & 33.87$\pm$26.93 & 0.566 \\
 & Gemini 3.1 Pro & 3.41$\pm$1.00 & 3.40$\pm$0.86 & 3.78$\pm$0.60 & 3.53$\pm$0.51 & \winner{0.952$\pm$0.127} & 0.262$\pm$0.096 & 15.30$\pm$12.88 & 0.611 \\
 & Qwen 3.6 Plus & 3.77$\pm$0.67 & 3.40$\pm$0.75 & 3.84$\pm$0.36 & \winner{3.67$\pm$0.38} & \winner{0.965$\pm$0.116} & 0.246$\pm$0.090 & 26.55$\pm$22.26 & 0.607 \\
 & KIMI K2.5 & 3.47$\pm$0.65 & 3.28$\pm$0.82 & 3.76$\pm$0.47 & 3.50$\pm$0.38 & \winner{0.963$\pm$0.112} & 0.236$\pm$0.086 & 23.82$\pm$17.87 & 0.605 \\
\midrule
\multirow{4}{*}{SeaSlides-Typst} & Claude Sonnet 4.6 & \best{4.19$\pm$0.67} & 3.91$\pm$0.47 & \winner{4.10$\pm$0.41} & \best{4.06$\pm$0.33} & \winner{0.981$\pm$0.077} & \winner{0.600$\pm$0.143} & 17.56$\pm$14.51 & \winner{0.792} \\
 & Gemini 3.1 Pro & \winner{3.53$\pm$0.96} & \winner{3.49$\pm$0.77} & \winner{3.98$\pm$0.48} & \winner{3.66$\pm$0.56} & \best{0.948$\pm$0.126} & \winner{0.574$\pm$0.171} & \best{8.83$\pm$6.16}\phantom{0} & \winner{0.807} \\
 & Qwen 3.6 Plus & 3.56$\pm$0.71 & 3.31$\pm$0.79 & \winner{4.09$\pm$0.51} & \best{3.65$\pm$0.47} & \best{0.931$\pm$0.139} & \winner{0.564$\pm$0.166} & \best{13.77$\pm$10.34} & \winner{0.766} \\
 & KIMI K2.5 & 3.49$\pm$0.82 & 3.53$\pm$0.74 & \winner{4.06$\pm$0.46} & \winner{3.69$\pm$0.43} & 0.881$\pm$0.164 & \winner{0.581$\pm$0.167} & \best{13.57$\pm$7.65}\phantom{0} & \winner{0.661} \\
\bottomrule
\end{tabular*}
\caption{Results on the combined 160-task setting. Content, Style, Readability, and Average use 1--5 scales; Constraint, CR, and Diversity use 0--1; Length is in thousands of characters. Within each model and excluding DeepPresenter, the best higher-is-better value is bold-underlined and values within 0.05 are bold. Length below 15K is bold, with the minimum also underlined.}
\label{tab:main}
\end{table*}

The comparison also makes the trade-off visible. PPT-Master often receives strong visual scores, but its generated SVG is long and difficult to inspect: Table~\ref{tab:main} reports 127.81K generated characters with Claude Sonnet 4.6, compared with 17.56K for SeaSlides-Typst. KCTV is compact and readable but substantially weaker on rich technical content. SeaSlides does not dominate every visual-style score; it combines competitive rendered quality with source that remains content-oriented and repairable.

\subsection{RQ2: Rich-Content Fidelity}

Table~\ref{tab:extended} reports mathematical rendering, code rendering, and visualization quality on tagged tasks 129--160. Each metric has its own five-point rubric and is judged from the complete rendered deck against the task and available source material. With Claude Sonnet 4.6, the three-metric macro-averages of SeaSlides-HTML and SeaSlides-Typst are 4.63 and 4.52, compared with 3.75 for PPT-Master and 1.43 for KCTV. SeaSlides also has the highest macro-average with Gemini 3.1 Pro and KIMI K2.5, whereas PPT-Master leads with Qwen 3.6 Plus. Thus, the two implementations handle specialized content effectively, but neither dominates every metric under every generation model.

\begin{table}[!ht]
\centering
\fontsize{8pt}{9pt}\selectfont
\setlength{\tabcolsep}{0.3pt}
\renewcommand{\arraystretch}{0.98}
\begin{tabular*}{\columnwidth}{@{\extracolsep{\fill}}llcccc@{}}
\toprule
\multicolumn{1}{c}{System} &
\multicolumn{1}{c}{Model} &
\multicolumn{1}{c}{Math} &
\multicolumn{1}{c}{Code} &
\multicolumn{1}{c}{Vis.} &
\multicolumn{1}{c@{}}{Avg.} \\
\midrule
\multirow{4}{*}{KCTV}
& C4.6 & 2.09$\pm$1.00 & 1.20$\pm$0.41 & 1.00$\pm$0.00 & 1.43 \\
& G3.1 & 1.96$\pm$0.82 & 1.25$\pm$0.44 & 1.00$\pm$0.00 & 1.40 \\
& Q3.6 & 1.70$\pm$0.76 & 1.15$\pm$0.37 & 1.04$\pm$0.20 & 1.30 \\
& K2.5 & 2.00$\pm$0.95 & 1.15$\pm$0.37 & 1.08$\pm$0.28 & 1.41 \\
\midrule
\multirow{4}{*}{DeepPresenter}
& C4.6 & 2.56$\pm$0.88 & 2.00$\pm$1.22 & 2.00$\pm$0.82 & 2.19 \\
& G3.1 & 2.67$\pm$0.58 & 4.00$\pm$0.00 & 2.33$\pm$0.58 & 3.00 \\
& Q3.6 & 2.71$\pm$0.76 & 3.33$\pm$1.21 & 2.50$\pm$1.29 & 2.85 \\
& K2.5 & 3.44$\pm$0.73 & 2.56$\pm$1.01 & 2.00$\pm$0.76 & 2.67 \\
\midrule
\multirow{4}{*}{PPT-Master}
& C4.6 & 3.74$\pm$0.62 & 3.75$\pm$0.55 & 3.76$\pm$0.60 & 3.75 \\
& G3.1 & 2.04$\pm$0.82 & 2.05$\pm$0.89 & 1.48$\pm$0.65 & 1.86 \\
& Q3.6 & \best{4.00$\pm$1.09} & \winner{4.00$\pm$0.79} & \winner{4.12$\pm$1.13} & \winner{4.04} \\
& K2.5 & 3.74$\pm$0.81 & 3.35$\pm$1.14 & \winner{3.76$\pm$0.88} & 3.62 \\
\midrule
\multirow{4}{*}{SeaSlides-HTML}
& C4.6 & \winner{4.83$\pm$0.39} & \winner{4.55$\pm$0.69} & \winner{4.52$\pm$0.51} & \winner{4.63} \\
& G3.1 & \best{3.48$\pm$0.73} & \best{3.45$\pm$0.94} & \best{3.12$\pm$0.88} & \best{3.35} \\
& Q3.6 & 3.87$\pm$1.18 & \best{3.90$\pm$0.85} & \best{3.96$\pm$0.89} & \best{3.91} \\
& K2.5 & \best{4.00$\pm$0.67} & \best{3.70$\pm$0.86} & \best{3.64$\pm$0.70} & \best{3.78} \\
\midrule
\multirow{4}{*}{SeaSlides-Typst}
& C4.6 & \best{4.61$\pm$0.50} & \winner{4.55$\pm$0.60} & \best{4.40$\pm$0.50} & \best{4.52} \\
& G3.1 & \winner{3.83$\pm$0.58} & \winner{3.85$\pm$0.59} & \winner{3.28$\pm$0.84} & \winner{3.65} \\
& Q3.6 & \winner{4.04$\pm$0.37} & 3.80$\pm$0.70 & 3.40$\pm$0.76 & 3.75 \\
& K2.5 & \winner{4.13$\pm$0.81} & \winner{3.75$\pm$1.02} & 3.60$\pm$1.22 & \winner{3.83} \\
\bottomrule
\end{tabular*}
\caption{Rich-content scores on tasks 129--160 (1--5; mean$\pm$standard deviation). Avg. is the macro-average of the three metrics and has no standard deviation because their subsets overlap. Model abbreviations follow Table~\ref{tab:main}: C4.6 (Claude), G3.1 (Gemini), Q3.6 (Qwen), and K2.5 (KIMI). DeepPresenter counts are in Appendix Table~\ref{tab:deeppresenter-results}. Per model, the best and second-best non-DeepPresenter values are bold-underlined and bold.}
\label{tab:extended}
\end{table}

\begin{figure*}[t]
    \centering
    \includegraphics[width=\textwidth]{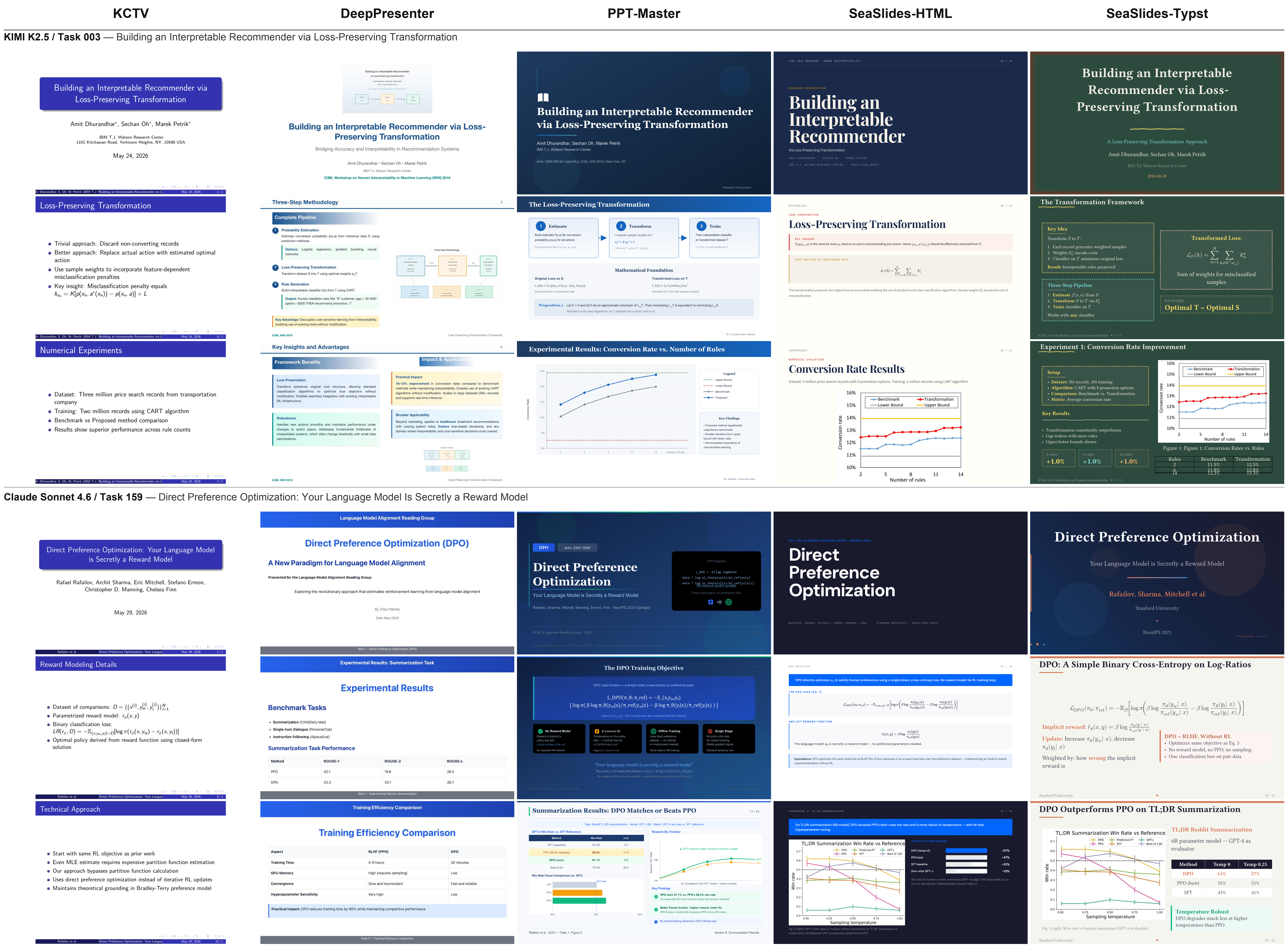}
    \caption{Task-matched qualitative comparison. Each row fixes the generation model and task: KIMI K2.5 on task 003, \emph{Building an Interpretable Recommender via Loss-Preserving Transformation}; and Claude Sonnet 4.6 on task 159, \emph{Direct Preference Optimization: Your Language Model Is Secretly a Reward Model}. Each column shows three slides from one system; the DeepPresenter cells use completed outputs.}
    \label{fig:qualitative-cases}
\end{figure*}

\subsection{RQ3: Ablating Semantic Abstraction and Feedback Stages}

Table~\ref{tab:ablation} evaluates SeaSlides on 40 tasks with failure-aware qualitative scores: outputs that fail to render or remain unusable receive low scores rather than disappearing from the denominator.

\begin{center}
\footnotesize
\setlength{\tabcolsep}{2.5pt}
\renewcommand{\arraystretch}{1.02}
\begin{tabular}{lccc}
\toprule
Configuration & System & Readability & Average \\
\midrule
Full system & HTML & \best{3.98$\pm$0.16} & \winner{4.05$\pm$0.40} \\
Full system & Typst & \best{4.13$\pm$0.40} & \winner{4.06$\pm$0.32} \\
\midrule
w/o Stage 1 & HTML & 3.80$\pm$0.85 & 3.68$\pm$1.03 \\
w/o Stage 1 & Typst & 2.68$\pm$1.64 & 2.54$\pm$1.52 \\
w/o Stage 2 & HTML & \best{3.93$\pm$0.27} & \best{3.96$\pm$0.52} \\
w/o Stage 2 & Typst & 4.03$\pm$0.36 & 3.77$\pm$0.47 \\
w/o Stage 3 & HTML & \winner{4.00$\pm$0.23} & 3.90$\pm$0.22 \\
w/o Stage 3 & Typst & \winner{4.23$\pm$0.42} & 3.95$\pm$0.29 \\
w/o TCS & HTML & 2.85$\pm$0.62 & 3.47$\pm$0.43 \\
w/o TCS & Typst & 2.95$\pm$0.45 & 3.23$\pm$0.52 \\
\bottomrule
\end{tabular}
\captionof{table}{Selected failure-aware ablation results on 40 tasks. Within each backend, the best value is bold-underlined and qualitative results within 0.10 are bold. Full metrics and row-specific valid-output counts are in Appendix Table~\ref{tab:ablation-full}.}
\label{tab:ablation}
\end{center}

Removing Stage 1 build diagnostics produces the largest cross-backend loss. The HTML and Typst averages fall to 3.68$\pm$1.03 and 2.54$\pm$1.52. Appendix Table~\ref{tab:ablation-full}(b) reports the corresponding valid-output counts and source metrics: 37/40 HTML runs and 21/40 Typst runs remain valid. On the 21 valid Typst outputs, constraint satisfaction, content ratio, and source length are 0.976$\pm$0.109, 0.584$\pm$0.146, and 10.65$\pm$5.33. Removing TCS most strongly harms source readability. Removing visual review has a smaller aggregate effect because earlier stages already catch many static defects, although perceptual problems still require rendered evidence.

\subsection{RQ4: Template Creation}

\paragraph{Template creation.}
We evaluate a harder setting in which the agent creates a task-specific template before authoring content. This setting adds design and component-definition decisions, so it is not expected to outperform curated template selection. Table~\ref{tab:template-summary} shows that template creation remains competitive but is not uniformly better: it improves SeaSlides-HTML with Qwen 3.6 Plus and KIMI K2.5, while SeaSlides-Typst with Claude Sonnet 4.6 remains close to its default counterpart. Appendix Table~\ref{tab:template-full} reports all qualitative, programmatic, source, and diversity metrics.

\begin{center}
\centering
\footnotesize
\setlength{\tabcolsep}{2.2pt}
\begin{tabular}{@{}lcccc@{}}
\toprule
& \multicolumn{2}{c}{SeaSlides-HTML} & \multicolumn{2}{c}{SeaSlides-Typst} \\
\cmidrule(lr){2-3}\cmidrule(l){4-5}
Model & Default & Create & Default & Create \\
\midrule
Claude Sonnet 4.6 & \textbf{4.05$\pm$0.40} & 3.84$\pm$0.60 & \textbf{4.06$\pm$0.32} & 4.03$\pm$0.40 \\
Gemini 3.1 Pro & \textbf{3.55$\pm$0.57} & 3.33$\pm$0.54 & \textbf{3.68$\pm$0.59} & 3.38$\pm$0.60 \\
Qwen 3.6 Plus & 3.56$\pm$0.41 & \textbf{3.68$\pm$0.40} & \textbf{3.62$\pm$0.40} & 3.45$\pm$0.52 \\
KIMI K2.5 & 3.47$\pm$0.40 & \textbf{3.49$\pm$0.56} & \textbf{3.74$\pm$0.42} & 3.65$\pm$0.49 \\
\bottomrule
\end{tabular}
\captionof{table}{Average qualitative scores on the 40-task template-generation experiment. Cells report mean$\pm$standard deviation; bold marks the higher value within each model and backend.}
\label{tab:template-summary}
\end{center}

\subsection{Case Study}
\label{sec:case-study}

Figure~\ref{fig:qualitative-cases} compares tasks 003 and 159. On task 003~\citep{dhurandhar2016interpretable}, KCTV and DeepPresenter reduce the source to bullets or a three-step summary; SeaSlides retains the transformed-loss equation, pipeline, and plot. PPT-Master uses underscore-style linear equations, while SeaSlides typesets subscripts, sums, and fractions. On task 159 (DPO)~\citep{rafailov2023dpo}, KCTV shows the reward-modeling loss; SeaSlides retains the policy/reference-policy log-ratio objective and temperature-sensitivity plot. The comparison concerns technical fidelity, not generic visual superiority.

\section{Limitations}

Our evaluation centers on static technical presentations. The current contracts cover common slide structures and specialized content, but not complex animation or tightly art-directed work. New rendering stacks still require backend-specific contracts and component implementations, whose maintenance cost we do not measure; HTML-to-PPTX conversion may simplify browser effects. Stylistic control remains a limitation: PPT-Master exposes a lower-level SVG interface and scores higher on Style in three of four models.

SeaSlidesBench-Rich is a 32-task capability stress test, not a representative sample of presentation domains.

The evidence is also limited: most qualitative results use GPT-5.5, source metrics only proxy editability, and the single-annotator study in Appendix~\ref{app:human-eval} is only a human sanity check. Each ablation condition is run once on 40 tasks, so failure rates and effect sizes may vary across trials. Broader domains, replication, and larger studies remain future work.

\section{Conclusion}

SeaSlides shows that a presentation agent can control narrative and evidence without authoring the renderer. Its semantic abstraction layer gives the model a backend-specific content vocabulary; templates and capability modules implement visual realization. Although HTML and Typst use different languages and rendering stacks, both yield substantially more readable, content-oriented source than SVG-heavy generation, and a SeaSlides backend achieves the highest rich-content macro-average under three of four models. Ablations identify build diagnostics as important to execution and template-content separation as important to editable source. Template-generation results show that the same boundary can support task-specific templates, although curated templates remain stronger in several settings. These findings support semantic abstraction as a practical design principle across presentation backends without requiring a shared language or renderer.

\clearpage
\bibliography{SeaSlides}
\clearpage

\appendix
\setcounter{secnumdepth}{1}
\raggedbottom
\setcounter{dbltopnumber}{1}
\renewcommand{\dbltopfraction}{0.95}
\renewcommand{\textfraction}{0.05}
\renewcommand{\dblfloatpagefraction}{0.75}
\makeatletter
\setlength{\@dblfptop}{0pt}
\setlength{\@dblfpbot}{0pt plus 1fil}
\makeatother

\section{Metric Definitions}
\label{app:metrics}

\begin{table*}[!t]
\centering
\textit{(a) Qualitative metrics.}
\par\vspace{1pt}
\small
\setlength{\tabcolsep}{9pt}
\begin{tabular}{llcccc}
\toprule
Configuration & System & Content & Style & Readability & Average \\
\midrule
Full system & HTML & \winner{4.18$\pm$0.84} & \best{4.00$\pm$0.78} & \best{3.98$\pm$0.16} & \winner{4.05$\pm$0.40} \\
Full system & Typst & \winner{4.15$\pm$0.77} & \winner{3.90$\pm$0.38} & \best{4.13$\pm$0.40} & \winner{4.06$\pm$0.32} \\
\midrule
w/o Stage 1 & HTML & 3.48$\pm$1.30 & 3.75$\pm$1.33 & 3.80$\pm$0.85 & 3.68$\pm$1.03 \\
w/o Stage 1 & Typst & 2.48$\pm$1.63 & 2.48$\pm$1.55 & 2.68$\pm$1.64 & 2.54$\pm$1.52 \\
w/o Stage 2 & HTML & 3.85$\pm$0.95 & \winner{4.10$\pm$0.84} & \best{3.93$\pm$0.27} & \best{3.96$\pm$0.52} \\
w/o Stage 2 & Typst & 3.98$\pm$0.80 & 3.30$\pm$0.72 & 4.03$\pm$0.36 & 3.77$\pm$0.47 \\
w/o Stage 3 & HTML & 3.90$\pm$0.50 & 3.80$\pm$0.41 & \winner{4.00$\pm$0.23} & 3.90$\pm$0.22 \\
w/o Stage 3 & Typst & 3.85$\pm$0.53 & 3.78$\pm$0.48 & \winner{4.23$\pm$0.42} & 3.95$\pm$0.29 \\
w/o TCS & HTML & 3.83$\pm$0.68 & 3.73$\pm$0.51 & 2.85$\pm$0.62 & 3.47$\pm$0.43 \\
w/o TCS & Typst & 3.70$\pm$0.88 & 3.05$\pm$0.71 & 2.95$\pm$0.45 & 3.23$\pm$0.52 \\
\bottomrule
\end{tabular}

\vspace{0pt}
\textit{(b) Programmatic, source, and diversity metrics.}
\par\vspace{1pt}
\footnotesize
\setlength{\tabcolsep}{5.2pt}
\begin{tabular}{llcccc@{\hspace{4pt}\vrule\hspace{4pt}}c}
\toprule
Configuration & System & Constraint & CR & Length & Diversity & $N$ \\
\midrule
Full system & HTML & \best{0.983$\pm$0.074} & \best{0.252$\pm$0.086} & 33.64$\pm$26.74 & \winner{0.460} & 40 \\
Full system & Typst & \best{0.975$\pm$0.089} & \best{0.593$\pm$0.123} & 16.74$\pm$14.08 & \winner{0.686} & 40 \\
\midrule
w/o Stage 1 & HTML & \winner{1.000$\pm$0.000} & \best{0.254$\pm$0.087} & \winner{27.95$\pm$19.83} & \best{0.448} & 37 \\
w/o Stage 1 & Typst & \best{0.976$\pm$0.109} & \best{0.584$\pm$0.146} & \winner{10.65$\pm$5.33} & 0.618 & 21 \\
w/o Stage 2 & HTML & \best{0.992$\pm$0.053} & \winner{0.265$\pm$0.090} & 30.61$\pm$23.28 & \best{0.433} & 40 \\
w/o Stage 2 & Typst & \winner{0.983$\pm$0.074} & \best{0.613$\pm$0.117} & 17.36$\pm$15.51 & 0.483 & 40 \\
w/o Stage 3 & HTML & \winner{1.000$\pm$0.000} & \best{0.258$\pm$0.082} & 29.90$\pm$21.14 & 0.316 & 40 \\
w/o Stage 3 & Typst & \winner{0.983$\pm$0.074} & \winner{0.615$\pm$0.130} & 17.12$\pm$13.59 & 0.564 & 40 \\
w/o TCS & HTML & \winner{1.000$\pm$0.000} & 0.186$\pm$0.106 & 44.45$\pm$28.20 & 0.401 & 40 \\
w/o TCS & Typst & 0.875$\pm$0.163 & 0.366$\pm$0.170 & 25.52$\pm$12.86 & 0.472 & 40 \\
\bottomrule
\end{tabular}
\caption{Full failure-aware ablation results on the 40-task subset. Panel (a) reports qualitative metrics; within each backend, the best value is bold-underlined and values within 0.10 are bold. In Panel (b), task-specification Constraint, CR, source length in thousand characters, and Diversity use the same valid-output subset whose size is shown by $N$. For 0--1 metrics, values within 0.05 of the best are bold; the shortest source is bold-underlined.}
\label{tab:ablation-full}
\end{table*}

\paragraph{Constraint satisfaction.}
Constraint satisfaction averages the applicable, explicitly specified presentation-level checks: page count, aspect ratio, and language. In the ablation programmatic panel, it uses the same valid-output subset as CR, source length, and Diversity, whose size is shown by $N$; the all-40-task aggregate is retained only as an internal audit field. Content-specific instructions are assessed by Content quality. For Typst projects, section-divider slides can be excluded from the effective slide count when the raw count also satisfies the task.

\paragraph{Source length.}
Source length is the length of relevant generated textual source in thousand characters. For systems with separated content and template files, copied templates and generated export artifacts are excluded where appropriate. For PPT-Master, source-length and content-ratio calculations use the generated SVG source before post-processing rather than the normalized final SVG export. In the ablation table, source length is aggregated only over the common valid-output subset with rendered images and no failure marker.

\paragraph{Content ratio.}
Content ratio estimates the proportion of generated source that represents content. Style declarations, colors, coordinates, repeated template code, generated export files, and copied scaffolding are excluded from the numerator. For TCS systems, the main denominator focuses on the LLM-authored content module; auxiliary debug metadata records the broader all-source ratio. In the ablation table, Constraint, CR, source length, and Diversity use the same valid-output subset for each row, whose size is shown by $N$.

\paragraph{Diversity.}
Diversity uses DINOv2 features and Vendi Score. The reported policy uses the first content slide in each project. The evaluation script also records first-slide, first-slides, mean-slides, and overview policies for debugging.

\paragraph{Content, style, and readability.}
Content measures faithfulness, completeness, and task relevance. Style measures visual design quality of rendered slides. Readability measures source readability and maintainability; it is evaluated from text inputs only and focuses on the model-authored content source unless a baseline lacks template-content separation. The evaluator returns an integer project-level score from 1 to 5 for each metric, and we compute means and sample standard deviations across projects.

\paragraph{Mathematical, code, and visualization rendering.}
These extended metrics are 1--5 integer scores on tagged tasks 129--160. Mathematical rendering assesses required-content coverage, semantic fidelity, proper mathematical typesetting, and legibility. Code rendering assesses the coverage and correctness of code and pseudocode together with syntax presentation and layout. Visualization assesses the data, labels, structure, and legibility of charts, tables, and diagrams. Missing required specialized content lowers the score rather than making the remaining slides appear error-free.

\section{LLM/VLM Evaluation Prompts}
\label{app:prompts}

We reproduce the evaluator templates used to generate the reported Content, Style, Readability, Math, Code, and Visualization scores. Each evaluation job concatenates the shared template below with exactly one metric-specific rubric. Runtime fields provide the task request and metadata; the accompanying input manifest identifies the rendered slide images, converted source material, or model-authored source files available to the evaluator. Content, Style, Math, Code, and Visualization use rendered slides, whereas Readability uses only the designated source files. Runtime values are shown as placeholders and Markdown emphasis markers are omitted for typesetting; the instructions and score anchors are unchanged. Constraint satisfaction, content ratio, source length, and Diversity are programmatic and therefore do not use an LLM/VLM prompt.

\subsection{Shared Prompt Template}

\begin{lstlisting}[style=seaslidesprompt]
# SeaSlidesBench {{metric}} Evaluation

Task ID: {{task_id}}
Dataset: {{dataset}}
Source: {{source}}
Tags: {{tags}}

## Task Prompt

{{task_prompt}}

## Required Output

Return one JSON object only:

{
  "score": <integer 1-5>,
  "rationale": "<brief reason>",
  "evidence": ["<specific observed evidence>"],
  "issues": ["<important problems, if any>"]
}
\end{lstlisting}

\subsection{Content Quality Prompt}

\begin{lstlisting}[style=seaslidesprompt]
## Rubric: Content Quality (1-5)

Score the rendered slides for completeness, accuracy, relevance, and logical
organization against the task prompt and any converted source Markdown.

- 5: Comprehensive, accurate, well organized, and faithful to the task/source.
- 4: Good coverage with minor omissions or small organization gaps.
- 3: Adequate but visibly shallow, incomplete, or uneven.
- 2: Major missing content, weak logic, or notable inaccuracies.
- 1: Mostly incomplete, irrelevant, or incorrect.
\end{lstlisting}

\subsection{Visual Style Prompt}

\begin{lstlisting}[style=seaslidesprompt]
## Rubric: Style Quality (1-5)

Score the visual design quality of the rendered slides.

- 5: Professional design, strong hierarchy, excellent typography, spacing,
  color, and polish.
- 4: Good design with minor visual issues.
- 3: Adequate but generic or uneven.
- 2: Poor visual quality, weak hierarchy, clutter, or bad spacing.
- 1: Visually broken or unusable.
\end{lstlisting}

\subsection{Source Readability Prompt}

\begin{lstlisting}[style=seaslidesprompt]
## Rubric: Readability / Human Editability (1-5)

Evaluate only the files listed in `read_focus_files` from inputs.json. Do not
use rendered images for this metric. For TCS systems, `read_focus_files` should
contain only the LLM-authored content module (for example main.typ or
content.html), not copied templates or generic wrappers. For non-TCS systems,
judge the effective LLM-generated source files.

Penalize hardcoded coordinates, inline visual styling, excessive boilerplate,
and content hidden in opaque layout code. Reward semantic components and clear
content structure.

- 5: Mostly semantic content; easy to edit without understanding the styling
  layer.
- 4: Predominantly semantic with occasional low-level layout details.
- 3: Mixed semantic and raw markup/styling.
- 2: Mostly raw layout/styling with content embedded in hard-to-edit code.
- 1: Opaque generated code; practical editing requires regeneration.
\end{lstlisting}

\subsection{Mathematical Rendering Prompt}

\begin{lstlisting}[style=seaslidesprompt]
## Rubric: Mathematical Rendering Quality (1-5)

Inspect every rendered slide listed in `image_inputs` and compare the visible
mathematics with the task prompt and converted source material. Score only the
mathematical content required by the task/source. Missing required mathematics
is a failure, not evidence that the remaining slides are error-free.

Evaluate all of the following:

1. Coverage: required equations, derivations, symbols, and mathematical
   structures are present.
2. Semantic fidelity: operators, variables, relations, fractions, roots,
   superscripts/subscripts, limits, matrices, cases, and multi-line derivations
   preserve the intended mathematical meaning.
3. Proper typesetting: structured mathematics is rendered as real mathematical
   notation rather than replaced by linear plain text or Unicode
   approximations. Isolated symbols such as alpha or O(n) in prose are
   acceptable; penalize substitutions only when they replace an equation or
   structured notation expected by the task.
4. Layout and legibility: delimiters scale correctly; baselines, alignment,
   line breaking, glyphs, spacing, and font size are readable; formulas are not
   clipped, overlapping, or missing.

Use these score anchors:

- 5: All required mathematics is present, semantically faithful, properly
  typeset, and consistently readable. No meaningful notation error, Unicode or
  plain-text equation substitute, clipping, or missing glyph is visible.
- 4: Nearly all required mathematics is correct and well typeset. Only minor
  local defects appear, such as slightly awkward spacing/alignment, one small
  omission, or a notation issue that does not change the meaning.
- 3: The core mathematical content is usable, but there are noticeable
  omissions or rendering problems. Examples include one meaning-affecting
  notation error, several minor errors, uneven legibility, or a limited mix of
  proper equations and plain-text/Unicode substitutes.
- 2: Major mathematical content is missing, incorrect, or hard to read. Errors
  are widespread or materially alter meaning; structured equations are often
  replaced by plain text/Unicode, or clipping/glyph failures affect much of the
  relevant content.
- 1: Required mathematics is absent or essentially unusable. Most formulas are
  missing, severely corrupted, replaced by inadequate text approximations, or
  unreadable.

Return exactly one integer from 1 to 5. A major meaning-changing formula error
prevents a score above 3. If most required mathematics is missing, substituted,
or unreadable, the score cannot exceed 2.
\end{lstlisting}

\subsection{Code and Pseudocode Rendering Prompt}

\begin{lstlisting}[style=seaslidesprompt]
## Rubric: Code and Pseudocode Rendering Quality (1-5)

Inspect every rendered slide listed in `image_inputs` and compare the visible
code or pseudocode with the task prompt and converted source material. Evaluate
only the code/pseudocode required by the task. Omission of required examples or
algorithm steps must lower the score.

Evaluate all applicable dimensions:

1. Coverage and fidelity: required code blocks, APIs, examples, algorithms,
   inputs/outputs, and key steps are present and match the task/source.
2. Code correctness: tokens, operators, identifiers, literals, indentation,
   nesting, and line breaks preserve valid or clearly intended code.
3. Pseudocode correctness: initialization, control flow, loop/branch structure,
   update order, termination, and returned values accurately express the
   intended algorithm. Do not judge pseudocode as though it must compile.
4. Syntax presentation: syntax highlighting, when used, is appropriate for the
   declared language and does not misclassify major token categories. Neutral
   monochrome code is not inherently incorrect, and absent line numbers are not
   a defect unless the task requires them.
5. Layout and legibility: monospace alignment, indentation, wrapping, font
   size, contrast, and spacing remain readable; no important line is clipped,
   overlapped, or visually merged with surrounding prose.

Use these score anchors:

- 5: All required code and pseudocode is present, faithful, logically correct,
  professionally formatted, and easy to read. Highlighting is accurate where
  used, and no important line or algorithm step is damaged.
- 4: The technical content is correct and readable with only minor local
  defects, such as a small highlighting mistake, one awkward wrap, a slight
  indentation inconsistency, or a nonessential omission.
- 3: The code/pseudocode remains understandable, but multiple presentation
  defects or limited correctness issues are visible. Examples include generic
  or inconsistent highlighting, several poor wraps, partial omission, or an
  ambiguous but still recognizable algorithm step.
- 2: Major code or algorithm content is incomplete, incorrect, or difficult to
  read. Broken indentation/control flow, widespread highlighting errors, severe
  wrapping/clipping, or substantial omissions undermine its use.
- 1: Required code/pseudocode is absent or essentially unusable. The visible
  content is mostly corrupted, unreadable, unrelated, or logically invalid.

Return exactly one integer from 1 to 5. A major code or algorithm error prevents
a score above 3. If most required code/pseudocode is missing, incorrect, or
unreadable, the score cannot exceed 2.
\end{lstlisting}

\subsection{Visualization Rendering Prompt}

\begin{lstlisting}[style=seaslidesprompt]
## Rubric: Visualization Rendering Quality (1-5)

Inspect every rendered slide listed in `image_inputs` and compare required
tables, charts, flowcharts, architecture diagrams, and other data graphics with
the task prompt and converted source material. Evaluate only visualization
types that the task actually requires. Decorative illustrations do not
substitute for a required explanatory or data visualization.

Evaluate all applicable dimensions:

1. Coverage and suitability: required visualizations are present and use a
   representation appropriate to the intended data or relationship.
2. Data and label fidelity: values, categories, text, units, axes, scales,
   legends, annotations, and table cells agree with the task/source and do not
   create a misleading interpretation.
3. Structural correctness: diagram nodes, edges, directions, grouping,
   hierarchy, sequence, and dependencies accurately express the intended
   relationships.
4. Legibility: labels and table text are readable; marks are distinguishable;
   alignment, contrast, and spacing are adequate; nothing important is clipped,
   overlapped, truncated, or hidden.
5. Explanatory value: the visualization communicates the required evidence or
   process without relying on decorative complexity. Visual attractiveness
   alone does not compensate for incorrect data or structure; overall
   aesthetics are evaluated separately by the Style metric.

Use these score anchors:

- 5: All required visualizations are present, faithful, structurally correct,
  and consistently readable. Data, labels, axes/legends, table cells, and
  diagram relationships contain no meaningful error.
- 4: Visualizations are correct and useful with only minor local defects, such
  as a small label omission, slightly crowded region, or noncritical alignment
  issue that does not change interpretation.
- 3: The main information is still understandable, but noticeable omissions or
  correctness/readability problems remain. Examples include incomplete labels,
  a partially ambiguous relationship, uneven table readability, or one material
  but localized data/structure issue.
- 2: Major visual content is missing, misleading, structurally wrong, or
  difficult to read. Incorrect scales/data mappings, wrong diagram connections,
  widespread clipping, or substantial omissions undermine interpretation.
- 1: Required visualizations are absent or essentially unusable. They are
  replaced by placeholders/prose, severely corrupted, unreadable, or materially
  misleading.

Return exactly one integer from 1 to 5. A materially misleading data mapping or
structural relationship prevents a score above 3. If most required visual
content is missing, wrong, or unreadable, the score cannot exceed 2.
\end{lstlisting}

\section{Paired System Comparisons}
\label{app:sem}

The main tables report standard deviation because variation across source documents, topics, and technical content types is part of the result. For direct system comparisons, the matched task IDs permit a more informative paired analysis.

The main system comparisons are paired: for a fixed generation model, competing systems are evaluated on the same 160 task IDs, with all qualitative judgments produced by the same evaluator. We therefore compute task-level differences between their average qualitative scores. For each project, that average is the mean of Content, Style, and Readability. Table~\ref{tab:paired-llm-avg} reports the mean difference $\Delta$, a 95\% paired bootstrap confidence interval over task differences using 5000 resamples, and the percentage of tasks with $\Delta>0$. Positive values favor the first system named in the comparison.

\begin{table*}[!t]
\centering
\footnotesize
\setlength{\tabcolsep}{4pt}
\renewcommand{\arraystretch}{0.94}
\begin{tabular*}{\textwidth}{@{\extracolsep{\fill}}p{0.29\textwidth}p{0.15\textwidth}ccp{0.14\textwidth}c@{}}
\toprule
Comparison & Model & $N$ & Mean $\Delta$ & 95\% bootstrap CI & $\Delta>0$ tasks \\
\midrule
SeaSlides-Typst -- PPT-Master & Claude Sonnet 4.6 & 160 & 0.496 & [0.419, 0.577] & 75.0\% \\
SeaSlides-Typst -- PPT-Master & Gemini 3.1 Pro & 160 & 0.923 & [0.790, 1.052] & 83.1\% \\
SeaSlides-Typst -- PPT-Master & Qwen 3.6 Plus & 160 & 0.335 & [0.238, 0.433] & 58.1\% \\
SeaSlides-Typst -- PPT-Master & KIMI K2.5 & 160 & 0.508 & [0.402, 0.615] & 66.9\% \\
SeaSlides-HTML -- PPT-Master & Claude Sonnet 4.6 & 160 & 0.498 & [0.417, 0.577] & 77.5\% \\
SeaSlides-HTML -- PPT-Master & Gemini 3.1 Pro & 160 & 0.790 & [0.671, 0.913] & 78.1\% \\
SeaSlides-HTML -- PPT-Master & Qwen 3.6 Plus & 160 & 0.352 & [0.273, 0.431] & 62.5\% \\
SeaSlides-HTML -- PPT-Master & KIMI K2.5 & 160 & 0.319 & [0.213, 0.431] & 55.0\% \\
SeaSlides-Typst -- KCTV & Claude Sonnet 4.6 & 160 & 0.921 & [0.842, 1.000] & 94.4\% \\
SeaSlides-Typst -- KCTV & Gemini 3.1 Pro & 160 & 0.660 & [0.558, 0.765] & 78.1\% \\
SeaSlides-Typst -- KCTV & Qwen 3.6 Plus & 160 & 0.704 & [0.606, 0.802] & 78.1\% \\
SeaSlides-Typst -- KCTV & KIMI K2.5 & 160 & 0.679 & [0.596, 0.767] & 84.4\% \\
SeaSlides-HTML -- KCTV & Claude Sonnet 4.6 & 160 & 0.923 & [0.844, 1.004] & 93.8\% \\
SeaSlides-HTML -- KCTV & Gemini 3.1 Pro & 160 & 0.527 & [0.431, 0.621] & 74.4\% \\
SeaSlides-HTML -- KCTV & Qwen 3.6 Plus & 160 & 0.721 & [0.633, 0.808] & 87.5\% \\
SeaSlides-HTML -- KCTV & KIMI K2.5 & 160 & 0.490 & [0.408, 0.573] & 70.0\% \\
SeaSlides-Typst -- SeaSlides-HTML & Claude Sonnet 4.6 & 160 & -0.002 & [-0.088, 0.081] & 31.3\% \\
SeaSlides-Typst -- SeaSlides-HTML & Gemini 3.1 Pro & 160 & 0.133 & [0.019, 0.246] & 51.3\% \\
SeaSlides-Typst -- SeaSlides-HTML & Qwen 3.6 Plus & 160 & -0.017 & [-0.106, 0.073] & 37.5\% \\
SeaSlides-Typst -- SeaSlides-HTML & KIMI K2.5 & 160 & 0.190 & [0.100, 0.279] & 54.4\% \\
\bottomrule
\end{tabular*}
\caption{Task-level paired differences in the main-table LLM average. Each row uses the 160 common tasks for the same generation model and evaluator. Positive $\Delta$ favors the first system in the comparison.}
\label{tab:paired-llm-avg}
\end{table*}

The paired results show that both SeaSlides backends are above PPT-Master and KCTV across all four models. The direct Typst--HTML gap is smaller: the Claude Sonnet 4.6 and Qwen 3.6 Plus intervals include zero, while Gemini 3.1 Pro and KIMI K2.5 give positive Typst--HTML intervals.

\section{Human Evaluation}
\label{app:human-eval}

As a limited human check on the GPT-5.5-based assessment, one computer-science graduate student rated the 40-task subset. It contains 160 presentations generated by Claude Sonnet 4.6 across KCTV, PPT-Master, SeaSlides-HTML, and SeaSlides-Typst. The evaluator scored Content, Style, and source Readability on 1--5 scales; system identities were hidden and presentation order was randomized. Table~\ref{tab:human-eval} reports the resulting means. Both SeaSlides variants rank above the baselines in overall score, but the study has only one annotator and should be read as a sanity check rather than an agreement analysis.

\begin{table}[!ht]
\centering
\footnotesize
\setlength{\tabcolsep}{4pt}
\renewcommand{\arraystretch}{1.02}
\begin{tabular}{lcccc}
\toprule
System & Content & Style & Read. & Avg. \\
\midrule
KCTV & 3.05 & 1.38 & 4.43 & 2.95 \\
PPT-Master & 4.33 & 4.03 & 1.50 & 3.28 \\
\midrule
SeaSlides-HTML & 4.08 & 3.65 & 3.65 & 3.79 \\
SeaSlides-Typst & 4.25 & 3.75 & 4.25 & 4.08 \\
\bottomrule
\end{tabular}
\caption{Human evaluation results on 40 tasks (160 presentations). Scores are means on 1--5 scales; Avg. is the mean of Content, Style, and Readability.}
\label{tab:human-eval}
\end{table}

\section{Dataset and Tags}
\label{app:dataset}

The combined evaluation contains 160 tasks, summarized in Table~\ref{tab:dataset-split}. Tasks 1--128 follow the UltraPresent validation setting used for standard presentation generation. Tasks 129--160 form SeaSlidesBench-Rich, the new benchmark designed to stress technical rendering and source-conditioned generation.

\begin{center}
\begin{minipage}{\columnwidth}
\centering
\footnotesize
\setlength{\tabcolsep}{3pt}
\begin{tabular}{@{}lrp{0.52\linewidth}@{}}
\toprule
Split & Tasks & Role \\
\midrule
UltraPresent-valid & 128 & Standard topic and source-conditioned generation \\
SeaSlidesBench-Rich & 32 & Extended math, code, pseudocode, and visualization stress tests \\
\bottomrule
\end{tabular}
\captionof{table}{Evaluation task sets used in the paper.}
\label{tab:dataset-split}
\end{minipage}
\end{center}

The extended split contains lecture notes, arXiv papers, and code-repository tasks. Tags are overlapping, because the same technical deck can require equations, code, algorithms, and figures. Table~\ref{tab:extended-tags} reports the tag counts. Mathematical rendering, code rendering, and visualization use separate denominators: 23 math-tagged tasks, 20 tasks with code or pseudocode, and 25 visualization-tagged tasks.

\begin{center}
\begin{minipage}{\columnwidth}
\centering
\footnotesize
\begin{tabular}{lr}
\toprule
Tag & Count \\
\midrule
Long document & 15 \\
Math & 23 \\
Code & 13 \\
Pseudocode & 11 \\
Code or pseudocode & 20 \\
Visualization & 25 \\
\midrule
Average tags per task & 2.72 \\
\bottomrule
\end{tabular}
\captionof{table}{Overlapping tags in SeaSlidesBench-Rich.}
\label{tab:extended-tags}
\end{minipage}
\end{center}

\section{DeepPresenter Completed-Output Subset}
\label{app:deeppresenter}

DeepPresenter did not produce a complete presentation in every run because its agent loop was unstable on some tasks. Accordingly, the qualitative scores in the main comparison are based only on the subset of runs that completed and produced final PPTX and PDF files. Table~\ref{tab:deeppresenter-results} compares this subset with the aggregate over all runs.

On the rich-content tasks, this completion criterion intersects the Math, Code, and Visualization tag subsets differently. The DeepPresenter sample counts used to compute the results in Table~\ref{tab:extended} are 9/5/7 for C4.6, 3/3/3 for G3.1, 7/6/4 for Q3.6, and 9/9/8 for K2.5. Other systems use the full tagged subsets of 23 Math tasks, 20 Code or Pseudocode tasks, and 25 Visualization tasks.

\begin{center}
\begin{minipage}{\columnwidth}
\centering
\fontsize{9pt}{9pt}\selectfont
\setlength{\tabcolsep}{0.4pt}
\begin{tabular*}{\columnwidth}{@{\extracolsep{\fill}}llcc@{}}
\toprule
Setting & Model & All runs & Completed output \\
\midrule
Main 160 & C4.6 & 2.61$\pm$0.67 (160) & 3.08$\pm$0.46 (37) \\
Main 160 & G3.1 & 2.56$\pm$0.71 (160) & 3.11$\pm$0.58 (39) \\
Main 160 & Q3.6 & 2.53$\pm$0.79 (160) & 3.21$\pm$0.51 (49) \\
Main 160 & K2.5 & 3.11$\pm$0.82 (160) & 3.59$\pm$0.51 (83) \\
Template 40 & C4.6 & 2.66$\pm$0.61 \,\,\,(40) & 3.11$\pm$0.47 \,\,\,(9) \\
Template 40 & G3.1 & 2.61$\pm$0.63 \,\,\,(40) & 3.07$\pm$0.66 (10) \\
Template 40 & Q3.6 & 2.55$\pm$0.76 \,\,\,(40) & 3.23$\pm$0.58 (13) \\
Template 40 & K2.5 & 3.04$\pm$0.90 \,\,\,(40) & 3.63$\pm$0.51 (19) \\
\bottomrule
\end{tabular*}
\captionof{table}{DeepPresenter average qualitative scores over all runs and the completed-output subset. Cells report mean$\pm$standard deviation over project-level averages, with $N$ in parentheses. Model abbreviations follow Table~\ref{tab:extended}.}
\label{tab:deeppresenter-results}
\end{minipage}
\end{center}

The main difficulty was recovery after an initial failure. Some failures originated in the execution environment rather than the reflection method itself, but upstream dependency and state-management errors could propagate through the agent loop to the final artifact. In one representative run, the design agent produced an incorrect 4:3 canvas and missing image references, then repeated incompatible checks, issued an edit call with empty arguments, and encountered page overflow. The run eventually recovered and produced a complete PPTX, showing that execution feedback can help while the recovery process remains brittle and loop-intensive. Similar chains in unsuccessful runs left partial exports, unmatched PPTX/PDF outputs, or no complete artifact.

\begin{table*}[p]
\centering
\captionsetup{font=small}
{\large\bfseries Full Template-Generation Results\par}
\vspace{0pt}
\textit{(a) Qualitative metrics.}
\par\vspace{2pt}
\fontsize{8pt}{9pt}\selectfont
\renewcommand{\arraystretch}{1.05}
\setlength{\aboverulesep}{0.25pt}
\setlength{\belowrulesep}{0.25pt}
\setlength{\tabcolsep}{5pt}
\begin{tabular}{@{}lllcccc@{\hspace{5pt}\vrule\hspace{5pt}}c@{\hspace{5pt}}}
\toprule
System & Setting & Model & Content & Style & Readability & Average & $N$ \\
\midrule
KCTV & default & Claude Sonnet 4.6 & 2.63$\pm$0.87 & 2.85$\pm$0.36 & 3.90$\pm$0.55 & 3.13$\pm$0.39 & 40 \\
KCTV & default & Gemini 3.1 Pro & 2.60$\pm$0.90 & 2.45$\pm$0.55 & 3.78$\pm$0.62 & 2.94$\pm$0.38 & 40 \\
KCTV & default & Qwen 3.6 Plus & 2.70$\pm$0.88 & 2.48$\pm$0.51 & 3.68$\pm$0.62 & 2.95$\pm$0.44 & 40 \\
KCTV & default & KIMI K2.5 & 2.70$\pm$0.76 & 2.55$\pm$0.55 & 3.80$\pm$0.41 & 3.02$\pm$0.37 & 40 \\
\midrule
DeepPresenter & default & Claude Sonnet 4.6 & 3.11$\pm$0.93 & 2.67$\pm$0.87 & 3.56$\pm$0.53 & 3.11$\pm$0.47 & 9 \\
DeepPresenter & default & Gemini 3.1 Pro & 3.40$\pm$0.84 & 2.50$\pm$0.85 & 3.30$\pm$0.95 & 3.07$\pm$0.66 & 10 \\
DeepPresenter & default & Qwen 3.6 Plus & 3.46$\pm$0.88 & 2.85$\pm$0.80 & 3.38$\pm$0.77 & 3.23$\pm$0.58 & 13 \\
DeepPresenter & default & KIMI K2.5 & 4.11$\pm$0.74 & 3.74$\pm$0.56 & 3.05$\pm$0.71 & 3.63$\pm$0.51 & 19 \\
\midrule
PPT-Master & default & Claude Sonnet 4.6 & \winner{4.20$\pm$0.94} & \winner{4.40$\pm$0.71} & 2.30$\pm$0.65 & 3.63$\pm$0.51 & 40 \\
PPT-Master & default & Gemini 3.1 Pro & 2.85$\pm$1.19 & 3.10$\pm$0.84 & 2.15$\pm$0.43 & 2.70$\pm$0.65 & 40 \\
PPT-Master & default & Qwen 3.6 Plus & \winner{3.85$\pm$0.92} & \winner{3.88$\pm$0.61} & 2.13$\pm$0.40 & 3.28$\pm$0.44 & 40 \\
PPT-Master & default & KIMI K2.5 & 3.48$\pm$1.34 & \winner{3.70$\pm$0.76} & 2.15$\pm$0.58 & 3.11$\pm$0.64 & 40 \\
\midrule
SeaSlides-HTML & default & Claude Sonnet 4.6 & \best{4.18$\pm$0.84} & 4.00$\pm$0.78 & 3.98$\pm$0.16 & \best{4.05$\pm$0.40} & 40 \\
SeaSlides-HTML & default & Gemini 3.1 Pro & 3.40$\pm$0.96 & \winner{3.43$\pm$0.90} & 3.83$\pm$0.71 & 3.55$\pm$0.57 & 40 \\
SeaSlides-HTML & default & Qwen 3.6 Plus & 3.53$\pm$0.75 & 3.33$\pm$0.69 & 3.83$\pm$0.38 & 3.56$\pm$0.41 & 40 \\
SeaSlides-HTML & default & KIMI K2.5 & 3.35$\pm$0.53 & 3.28$\pm$0.93 & 3.78$\pm$0.48 & 3.47$\pm$0.40 & 40 \\
\midrule
SeaSlides-Typst & default & Claude Sonnet 4.6 & \best{4.15$\pm$0.77} & 3.90$\pm$0.38 & \best{4.13$\pm$0.40} & \winner{4.06$\pm$0.32} & 40 \\
SeaSlides-Typst & default & Gemini 3.1 Pro & \winner{3.53$\pm$1.01} & \winner{3.43$\pm$0.81} & \winner{4.08$\pm$0.42} & \winner{3.68$\pm$0.59} & 40 \\
SeaSlides-Typst & default & Qwen 3.6 Plus & 3.35$\pm$0.58 & 3.40$\pm$0.74 & \winner{4.10$\pm$0.50} & \best{3.62$\pm$0.40} & 40 \\
SeaSlides-Typst & default & KIMI K2.5 & 3.48$\pm$0.85 & 3.58$\pm$0.75 & \winner{4.18$\pm$0.38} & \winner{3.74$\pm$0.42} & 40 \\
\midrule
SeaSlides-HTML & create & Claude Sonnet 4.6 & 3.85$\pm$1.05 & 3.75$\pm$0.95 & 3.93$\pm$0.27 & 3.84$\pm$0.60 & 40 \\
SeaSlides-HTML & create & Gemini 3.1 Pro & 3.23$\pm$1.00 & 2.95$\pm$0.85 & 3.80$\pm$0.46 & 3.33$\pm$0.54 & 40 \\
SeaSlides-HTML & create & Qwen 3.6 Plus & 3.58$\pm$0.78 & 3.70$\pm$0.72 & 3.75$\pm$0.49 & \winner{3.68$\pm$0.40} & 40 \\
SeaSlides-HTML & create & KIMI K2.5 & 3.15$\pm$1.33 & \best{3.65$\pm$0.70} & 3.68$\pm$0.57 & 3.49$\pm$0.56 & 40 \\
\midrule
SeaSlides-Typst & create & Claude Sonnet 4.6 & 4.05$\pm$0.81 & 3.85$\pm$0.70 & \winner{4.18$\pm$0.45} & \best{4.03$\pm$0.40} & 40 \\
SeaSlides-Typst & create & Gemini 3.1 Pro & 3.28$\pm$0.99 & 2.88$\pm$0.82 & \best{4.00$\pm$0.60} & 3.38$\pm$0.60 & 40 \\
SeaSlides-Typst & create & Qwen 3.6 Plus & 3.30$\pm$0.91 & 3.20$\pm$0.85 & 3.85$\pm$0.70 & 3.45$\pm$0.52 & 40 \\
SeaSlides-Typst & create & KIMI K2.5 & \winner{3.63$\pm$0.77} & 3.43$\pm$0.78 & 3.90$\pm$0.30 & \best{3.65$\pm$0.49} & 40 \\
\bottomrule
\end{tabular}

\vspace{0pt}
\textit{(b) Programmatic, source, and diversity metrics.}
\par\vspace{1pt}
\setlength{\tabcolsep}{5.0pt}
\begin{tabular}{@{}lllccrc@{\hspace{5pt}\vrule\hspace{5pt}}c@{\hspace{5pt}}}
\toprule
System & Setting & Model & Constraint & CR & Length & Diversity & $N$ \\
\midrule
KCTV & default & Claude Sonnet 4.6 & 0.504$\pm$0.208 & 0.444$\pm$0.174 & \winner{3.61$\pm$1.40} & 0.225 & 40 \\
KCTV & default & Gemini 3.1 Pro & 0.504$\pm$0.260 & 0.460$\pm$0.159 & \winner{3.79$\pm$1.41} & 0.264 & 40 \\
KCTV & default & Qwen 3.6 Plus & 0.521$\pm$0.221 & 0.457$\pm$0.170 & \winner{3.58$\pm$1.47} & 0.240 & 40 \\
KCTV & default & KIMI K2.5 & 0.517$\pm$0.247 & 0.454$\pm$0.162 & \winner{3.58$\pm$1.49} & 0.267 & 40 \\
\midrule
DeepPresenter & default & Claude Sonnet 4.6 & 0.963$\pm$0.111 & 0.163$\pm$0.116 & 26.05$\pm$8.46 & 0.462 & 9 \\
DeepPresenter & default & Gemini 3.1 Pro & 1.000$\pm$0.000 & 0.179$\pm$0.096 & 22.91$\pm$7.62 & 0.363 & 10 \\
DeepPresenter & default & Qwen 3.6 Plus & 0.974$\pm$0.092 & 0.202$\pm$0.114 & 23.23$\pm$8.05 & 0.395 & 13 \\
DeepPresenter & default & KIMI K2.5 & 0.974$\pm$0.115 & 0.177$\pm$0.098 & 38.47$\pm$21.66 & 0.379 & 19 \\
\midrule
PPT-Master & default & Claude Sonnet 4.6 & 0.892$\pm$0.183 & 0.077$\pm$0.037 & 130.19$\pm$92.79 & 0.393 & 40 \\
PPT-Master & default & Gemini 3.1 Pro & 0.842$\pm$0.244 & 0.107$\pm$0.057 & 20.77$\pm$12.53 & 0.452 & 40 \\
PPT-Master & default & Qwen 3.6 Plus & 0.892$\pm$0.212 & 0.076$\pm$0.033 & 89.06$\pm$59.51 & 0.432 & 40 \\
PPT-Master & default & KIMI K2.5 & 0.842$\pm$0.256 & 0.078$\pm$0.040 & 81.68$\pm$44.88 & 0.358 & 40 \\
\midrule
SeaSlides-HTML & default & Claude Sonnet 4.6 & \best{0.983$\pm$0.074} & 0.252$\pm$0.086 & 33.64$\pm$26.74 & 0.460 & 40 \\
SeaSlides-HTML & default & Gemini 3.1 Pro & \winner{0.975$\pm$0.089} & 0.256$\pm$0.108 & \best{14.54$\pm$11.10} & 0.486 & 40 \\
SeaSlides-HTML & default & Qwen 3.6 Plus & \best{0.975$\pm$0.089} & 0.238$\pm$0.091 & 26.69$\pm$22.31 & 0.480 & 40 \\
SeaSlides-HTML & default & KIMI K2.5 & \winner{0.967$\pm$0.126} & 0.232$\pm$0.093 & 22.08$\pm$15.34 & 0.522 & 40 \\
\midrule
SeaSlides-Typst & default & Claude Sonnet 4.6 & \best{0.975$\pm$0.089} & \winner{0.593$\pm$0.123} & 16.74$\pm$14.08 & \best{0.686} & 40 \\
SeaSlides-Typst & default & Gemini 3.1 Pro & 0.921$\pm$0.151 & \winner{0.559$\pm$0.166} & \best{9.27$\pm$7.79} & \winner{0.668} & 40 \\
SeaSlides-Typst & default & Qwen 3.6 Plus & 0.925$\pm$0.141 & \winner{0.566$\pm$0.178} & \best{13.25$\pm$9.48} & \winner{0.698} & 40 \\
SeaSlides-Typst & default & KIMI K2.5 & 0.867$\pm$0.165 & \winner{0.571$\pm$0.156} & \best{13.19$\pm$7.26} & 0.569 & 40 \\
\midrule
SeaSlides-HTML & create & Claude Sonnet 4.6 & \winner{0.992$\pm$0.053} & 0.114$\pm$0.055 & 62.37$\pm$29.44 & 0.547 & 40 \\
SeaSlides-HTML & create & Gemini 3.1 Pro & \winner{0.975$\pm$0.089} & 0.160$\pm$0.096 & \best{13.61$\pm$6.84} & 0.613 & 40 \\
SeaSlides-HTML & create & Qwen 3.6 Plus & \winner{0.983$\pm$0.074} & 0.114$\pm$0.067 & 37.80$\pm$14.19 & 0.576 & 40 \\
SeaSlides-HTML & create & KIMI K2.5 & 0.867$\pm$0.221 & 0.104$\pm$0.053 & 39.64$\pm$18.19 & 0.564 & 40 \\
\midrule
SeaSlides-Typst & create & Claude Sonnet 4.6 & \winner{0.992$\pm$0.053} & 0.218$\pm$0.124 & 38.66$\pm$15.04 & \winner{0.687} & 40 \\
SeaSlides-Typst & create & Gemini 3.1 Pro & \best{0.925$\pm$0.155} & 0.280$\pm$0.164 & \best{12.04$\pm$4.54} & \best{0.665} & 40 \\
SeaSlides-Typst & create & Qwen 3.6 Plus & 0.904$\pm$0.160 & 0.240$\pm$0.144 & 22.76$\pm$8.90 & \best{0.693} & 40 \\
SeaSlides-Typst & create & KIMI K2.5 & 0.896$\pm$0.163 & 0.205$\pm$0.100 & 24.95$\pm$11.30 & \winner{0.729} & 40 \\
\bottomrule
\end{tabular}
\caption{Full template-generation results on the 40-task subset. Panel (a) reports qualitative metrics; Panel (b) reports programmatic, source, and diversity metrics. Each $N$ is the panel-specific sample count, and DeepPresenter follows the completed-output definition in Appendix Table~\ref{tab:deeppresenter-results}. Best non-DeepPresenter values are bold-underlined; qualitative values within 0.10 and 0--1 values within 0.05 are bold. Lengths below 15K are bold, with the minimum underlined.}
\label{tab:template-full}
\end{table*}

\FloatBarrier

\section{Implementation Lineage and Authoring Examples}
\label{app:implementation}

SeaSlides is developed from two open-source presentation skills, and we do not claim their inherited infrastructure as a contribution. The project lifecycle, source-conversion utilities, Strategist--Image Generator--Executor workflow, and parts of the asset and export tooling originate from PPT-Master~\citep{he2026pptmaster}. The HTML implementation additionally starts from the single-file web-deck templates and visual vocabulary of guizang-ppt-skill~\citep{op7418guizang2026}.

SeaSlides-HTML changes the inherited single-file workflow into an enforced contract among \texttt{content.html}, \texttt{template.css}, and \texttt{index-template.html}. It adds reusable semantic operations, settles the browser DOM before SVG serialization, and supports staged review together with PNG and editable PPTX export.

SeaSlides-Typst implements the rendering backend in Typst and Touying rather than porting the HTML or SVG source. It separates \texttt{main.typ} from \texttt{template.typ}, supplies a reusable theme and component library, and uses compiler diagnostics for local repair before PDF, PNG, or PPTX delivery. This work also adds SeaSlidesBench-Rich, the TCS checks, capability-module evaluation, and the benchmark harness. Figure~\ref{fig:wysiwyg-editor} and Listings~\ref{lst:typst-authoring-example}--\ref{lst:html-authoring-example} illustrate the resulting authoring boundary: both listings express the same slide intent without embedding its visual implementation.

\noindent
\begin{minipage}[t]{\columnwidth}
\vspace{0pt}
\begin{lstlisting}[style=seaslidescode,language=Typst]
#import "template.typ": *
#import "@preview/mitex:0.2.7": *

== Attention Mechanism

#concept-card([Scaled dot-product], [
  #mitex(`
    \text{Attention}(Q,K,V)
      = \text{softmax}\left(\frac{QK^\top}{\sqrt{d_k}}\right)V
  `)
])

#cols[
  - Multi-head: #mi(`h`) parallel heads
  - Complexity: #mi(`O(n^2 \cdot d)`)
][
  ```python
  scores = q @ k.T / math.sqrt(d_k)
  attn = F.softmax(scores, dim=-1)
  out = attn @ v
  ```
]

#speaker-note[Explain why scaling stabilizes the logits.]
\end{lstlisting}
\captionsetup{justification=raggedright,singlelinecheck=false}
\captionof{listing}{SeaSlides-Typst: model-authored \texttt{main.typ}.}
\label{lst:typst-authoring-example}
\end{minipage}

\begin{center}
\begin{minipage}{0.82\columnwidth}
\centering
\includegraphics[width=\linewidth]{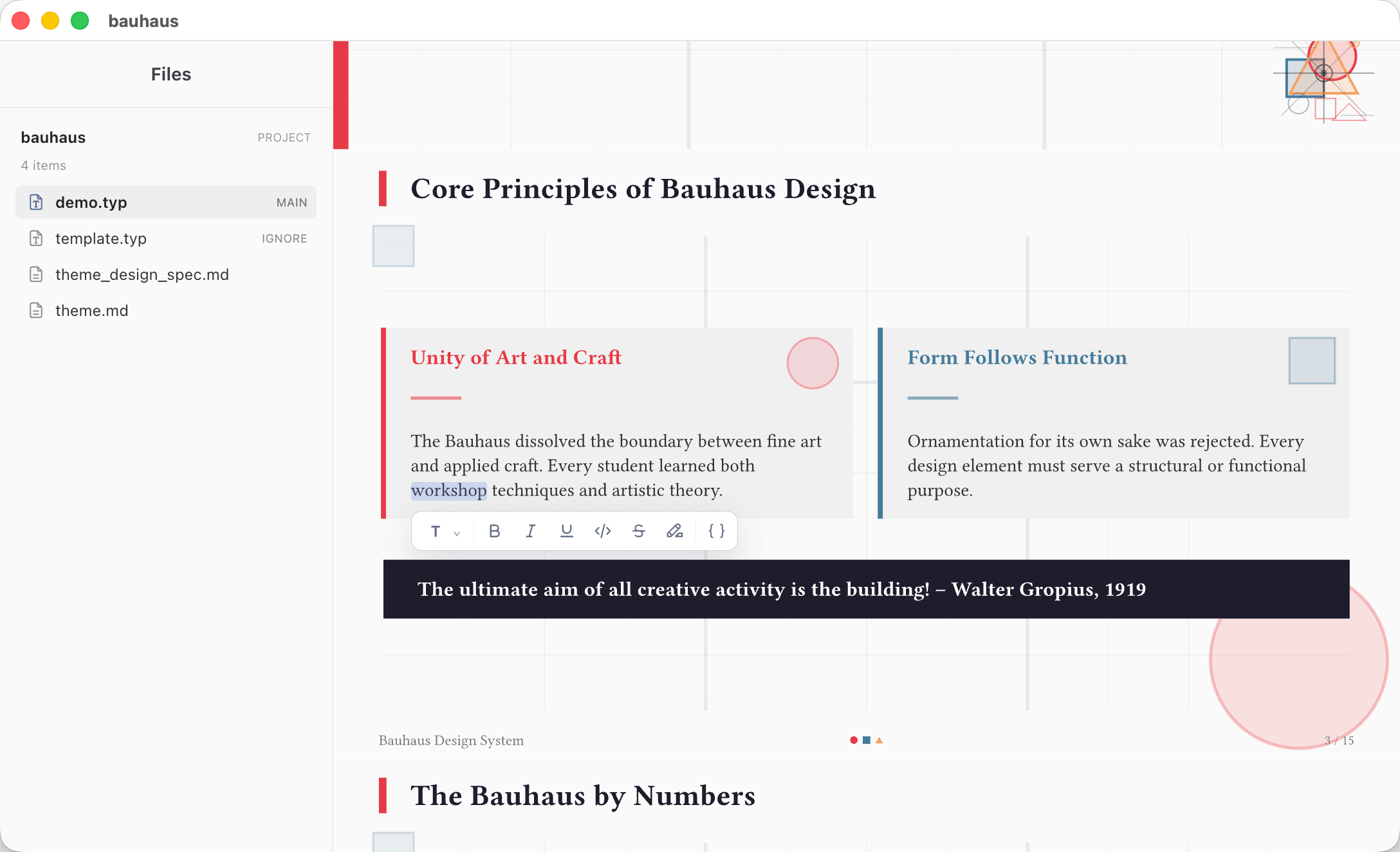}
\captionsetup{justification=raggedright,singlelinecheck=false}
\captionof{figure}{WYSIWYG editing interface for SeaSlides-Typst, showing project files, rendered-slide preview, and inline content editing.}
\label{fig:wysiwyg-editor}
\end{minipage}
\end{center}

\noindent
\begin{minipage}[t]{\columnwidth}
\vspace{0pt}
\begin{lstlisting}[style=seaslidescode,language=HTML]
<section class="slide" data-layout="IT06"
  data-notes="Explain why scaling stabilizes the logits.">
  <header class="slide-header"><h2>Attention Mechanism</h2></header>

  <div class="concept-card">
    <h3>Scaled dot-product</h3>
    <div class="math-block">\[
      \operatorname{Attention}(Q,K,V)
      = \operatorname{softmax}\!\left(\frac{QK^\top}{\sqrt{d_k}}\right)V
    \]</div>
  </div>

  <div class="cols cols-2">
    <ul><li>Multi-head: \(h\) parallel heads</li>
        <li>Complexity: \(O(n^2 d)\)</li></ul>
    <pre class="code-block"><code class="language-python">
scores = q @ k.T / math.sqrt(d_k)
attn = F.softmax(scores, dim=-1)
out = attn @ v
    </code></pre>
  </div>
</section>
\end{lstlisting}
\captionsetup{justification=raggedright,singlelinecheck=false}
\captionof{listing}{SeaSlides-HTML: model-authored \texttt{content.html}.}
\label{lst:html-authoring-example}
\end{minipage}

\end{document}